\documentclass[sigconf]{acmart}
\usepackage{enumitem}
\usepackage{multirow}
\usepackage{booktabs}
\usepackage{verbatim}
\usepackage{graphicx}

\AtBeginDocument{%
  \providecommand\BibTeX{{%
    \normalfont B\kern-0.5em{\scshape i\kern-0.25em b}\kern-0.8em\TeX}}}

\setcopyright{acmcopyright}
\copyrightyear{2023}
\acmYear{2023}
\acmDOI{XXXXXXX.XXXXXXX}

\acmConference[Conference acronym 'XX]{ }{June 03--05,
  2023}{Woodstock, NY}
\acmPrice{15.00}
\acmISBN{978-1-4503-XXXX-X/18/06}

\begin{document}

\sloppy
\title{A Deep Behavior Path Matching Network for Click-Through Rate Prediction
}

\author{Jian Dong$^1$, Yisong Yu$^{2,3}$, Yapeng Zhang$^1$, Yimin Lv$^{2,3}$, Shuli Wang$^1$, Beihong Jin$^{2,3*}$,   Yongkang Wang$^{1*}$, Xingxing Wang$^1$, Dong Wang$^1$}
\affiliation{
  \institution{$^1$Meituan}
  \institution{$^2$Institute of Software, Chinese Academy of Sciences}
  \institution{$^3$University of Chinese Academy of Sciences, Beijing, China}
  \city{}
  \country{}
  }
\email{Dongjian03@meituan.com, Beihong@iscas.ac.cn, Wangyongkang03@meituan.com}

\renewcommand{\shortauthors}{J. Dong and Y. Yu, et al.}

\begin{abstract}
  User behaviors on an e-commerce app not only contain different kinds of feedback on items but also sometimes imply the cognitive clue of the user's decision-making. For understanding the psychological procedure behind user decisions, we present the behavior path and propose to match the user's current behavior path with historical behavior paths to predict user behaviors on the app. Further, we design a deep neural network for behavior path matching and solve three difficulties in modeling behavior paths: sparsity, noise interference, and accurate matching of behavior paths. In particular, we leverage contrastive learning to augment user behavior paths, provide behavior path self-activation to alleviate the effect of noise, and adopt a two-level matching mechanism to identify the most appropriate candidate. Our model shows excellent performance on two real-world datasets, outperforming the state-of-the-art CTR model. Moreover, our model has been deployed on the Meituan food delivery platform and has accumulated 1.6\% improvement in CTR and 1.8\% improvement in advertising revenue. 

\end{abstract}




\maketitle

\section{Introduction}
Meituan Takeout APP is an app for catering and retail. Through the app, users can browse and choose the POIs (Points Of Interest, such as restaurants, food stores, and cafes) and place orders for food that will be fast delivered to users. The app is expected to understand the psychology behind user decisions and push relevant candidates to users, thus increasing the click-through rate (CTR) and further transaction volume and advertising revenue. 

We note that user behaviors on an app are an important manifestation of the user's decision-making psychology. However, although some existing models for CTR prediction have analyzed user behaviors, from the perspective of long sequences or multiple kinds of behaviors, they adopt the point-to-point activation of the candidate and individual behavior in the historical behavior sequence, without taking into consideration the influence of sequential behaviors which contain the user decision-making trail. Therefore, for the behavior of clicking a target POI, we view the user's sequential behaviors before that, including browsing the POIs, placing an order and etc., as a behavior path. By observing the historical data on Meituan Takeout APP, we find that there is a close correlation between the behavior path and the click behavior.

The above observations motivate us to develop a model that performs the behavior path matching to predict the user’s next click. The core idea is to learn latent factors related to decision-making psychology from the user behavior paths for generating their embeddings. Having in hand the embeddings from behavior paths, the model will perform the matching between historical behavior paths and the current behavior path and estimate the CTRs of candidates. 

However, it is challenging to model user behavior paths, because there exist three difficulties: the sparsity of behavior paths, the interference of noise in behavior paths, and the exact matching between behavior paths. Firstly, for a single user, the interactions between the user and the app are not much, which leads to the difficulty in capturing all behavior patterns of the user. For dealing with the sparsity of behavior paths, we leverage contrastive learning to augment the positives of user behavior paths and optimize the learning of the user behavior paths. Secondly, there is a lot of noise in the user behavior paths. For example, a user clicks on a POI due to its attractive cover but immediately returns as soon as the user feels he/she does not like it. Such behavior actually becomes the noise in the path. For reducing the impact of noise, we build a dynamic activation network to focus on several main behaviors of the path. 
Compared to equally treating all the behaviors in a path, the dynamic activation is more effective and efficient, since some behaviors do have a more obvious effect on subsequent behaviors. Finally, we propose a two-level matching mechanism. In the first level, for the current path, we calculate the activation weight of each historical behavior path and then choose the top-k most similar historical paths. In the second level, given the candidate and the chosen paths, we calculate the activation weights of the click behaviors which follow the chosen paths for CTR prediction.

Our main contributions are summarized as follows.
\begin{itemize}[leftmargin=*]
\item We are the first to introduce user behavior path matching into the industrial CTR prediction. We identify the challenges of modeling behavior paths, i.e., the sparsity, noise, and matching problems of behavior paths. 
\item We propose a Deep Behavior Path Matching Network (DBPMaN) to predict CTRs, which augments behavior paths, provides behavior path self-activation, 
and performs a two-level matching (at the level of behavior paths first and then at the level of click behaviors) for CTR prediction.
\item We conduct offline experiments on two real-world datasets of different scales and the online A/B test in Meituan advertising. The experimental results show DBPMaN is effective and achieves state-of-the-art performance.
\end{itemize}

\section{Related Work}
CTR prediction, as a kernel part of the recommender system, has been a hot topic concerned by industry and academia \cite{zhang2021deep}.

The classic solution to CTR prediction is to learn feature interactions, where DeepFM \cite{guo2017deepfm}, xDeepFM \cite{lian2018xdeepfm}, and ONN \cite{yang2020onn} are early representative deep neural network models and CAN \cite{bian2022can} has the state-of-the-art performance among current open source CTR models. 

Recently, sequential behavior modeling becomes a new driving force for CTR prediction. The granularity of modeled behaviors ranges from the single behavior (e.g., DIN \cite{zhou2018din}) to multiple kinds of behaviors (e.g., FeedRec \cite{wu2022feedrec}), from short sequences (e.g., DIEN \cite{zhou2019dien}, DSIN \cite{feng2019dsin}) to ultra-long sequences (e.g., MIMN \cite{pi2019mimn}, SIM \cite{pi2020sim}, ETA \cite{chen2021eta}). These models aim to capture user interests \cite{zhou2018din,zhou2019dien,feng2019dsin,zheng2022clsr} or intention \cite{li2019gin} and often adopt the point-to-point activation method, whose input only contains a single kind of behaviors such as click, to estimate the tendency of user interests/intention towards candidates from a probabilistic perspective. They are proven to have continued to improve the accuracy of CTR prediction. In addition, with the great success of Transformer \cite{vaswani2017transformer} and BERT \cite{devlin2018bert} in the field of NLP, it has been introduced into recommender systems to achieve different recall tasks \cite{kang2018self, sun2019bert4rec} or CTR predication tasks \cite{chen2019bst}. 

Compared to the above work, our work emphasizes behavior paths that imply the decision-making signs and employs behavior paths as the evidence base of CTR prediction.

\section{Our Model}
\subsection{Overview}
Firstly, we give the following definitions used in the paper.

\vspace{+1pt}
\noindent \textbf{Definition 1. (User Behavior Sequence)}  Let $\mathcal{U}$ denote a user set. For the user $u \in \mathcal{U}$, his/her behavior sequence is composed of his/her behaviors, sorted by occurrence time and denoted by $s = [b_1, ..., b_i, ..., b _T]$, where $b_i$ is the $i$-th behavior and T is the length of the behavior sequence. In our scenario, this sequence includes user behaviors during the past year. Each behavior includes the id of the interacted item, the behavior type, the interval between the occurrence time and the current time and the relative position in the sequence, etc. In our scenario, there exist three behavior types: click, impression, and order. 

\vspace{+1pt}
\noindent \textbf{Definition 2. (User Click Sequence)} In the user behavior sequence $s$, there exist a large number of click behaviors. Thus we can form a click sequence $s^c$ from $s$: $s^c= [b_1^c,...,b_i^c,...,b_t^c]$,  where $t$ denotes the length of the click sequence.

\vspace{+1pt}
\noindent \textbf{Definition 3. (User Behavior Path)} For the $i$-th click behavior $b_i^c$ in $s^c$, let $b_{m(i)}$
denote the corresponding behavior in the behavior sequence $s$, where $m(i)$ denotes the position in $s$ where the $i$-th click behavior occurs. Then, the user behavior path with respect to the click behavior $b_i^c$, denoted by $p_i$, is the subsequence $[b_{m(i)-l},...,b_{m(i)-2}, b_{m(i)-1}]$ in $s$, where $l$ is the preset length of the behavior path.

\begin{figure}[htbp]
  \centering
  \includegraphics[width=\linewidth]{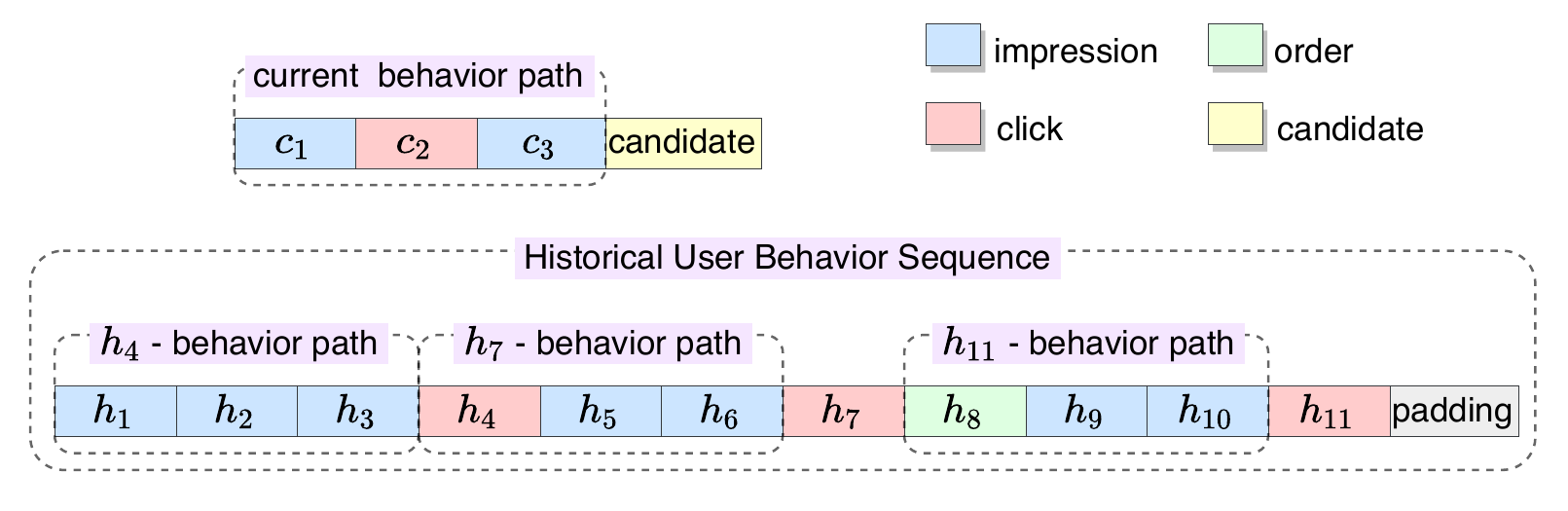}
  \caption{Example of user behavior paths where $l$=3.}
  \label{behavior path}
\end{figure}

From the definition of the user behavior path, we can obviously find that the click behavior and the behavior path are one-to-one correspondence. Fig. \ref{behavior path} gives an example of user behavior paths. In the historical user behavior sequence, there are three user behavior paths whose length is preset to 3: one w.r.t. $h_{4}$, one w.r.t. $h_{7}$, and one w.r.t. $h_{11}$.

\vspace{+1pt}
\noindent \textbf{Definition 4. (Behavior Path Sequence)}  For all click behaviors in $s$, we can obtain their corresponding behavior paths, respectively, thus forming a behavior path sequence $ P=[p_1,...,p_i,...,p_t]$, where $p_i$ is the user behavior path w.r.t. the click behavior $b_i^c$.

\vspace{+1pt}
Secondly, we give the composition, structure and process of the DBPMaN model.

\begin{figure}[tbp]
	\centering
	\includegraphics[width=\linewidth]{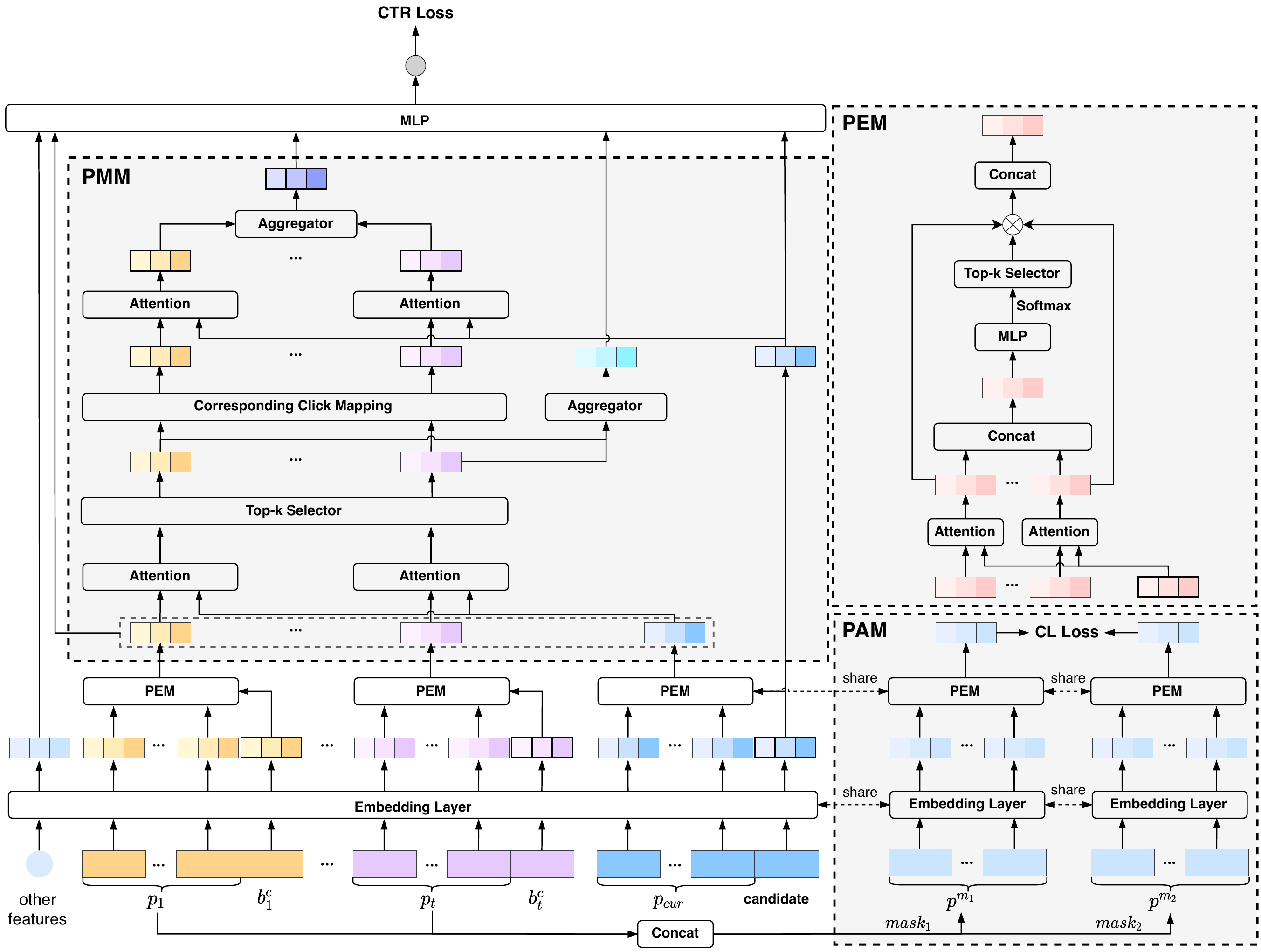}
	\caption{Structure of DBPMaN. (1) PEM enhances the path representations. (2) PMM matches the paths. (3) PAM augments the paths.}
	\label{model architecture}
\vspace{-15pt}
\end{figure}

DBPMaN consists of an embedding layer and three modules, i.e., Path Enhancing Module (PEM), Path Matching Module (PMM), and Path Augmenting Module (PAM), whose structure is shown in Fig. \ref{model architecture}. 

DBPMaN takes multiple features as input, where the chosen features are from (1) item profile, including item id and its side information (e.g., category, location, rating, etc.); (2) user profile, including user id and his/her side information (e.g., age, gender, city, etc.); (3) behaviors in the user behavior sequence. These features are fed into the embedding layer. For all features, we can obtain their embeddings by looking up the embedding tables. Then, we use sum pooling on different kinds of features to calculate the embedding of user behavior sequence $\mathbf{s} = [\mathbf{e}_1,...,\mathbf{e}_i,...,\mathbf{e}_T]$ , user embedding $\mathbf{e}^u$ and the embedding of candidate item $\mathbf{e}^{ct}$.

The behaviors in a user behavior path tend to contribute differently to the corresponding next click behavior, i.e., the click behavior which follows the user behavior path. PEM is expected to mine this information to learn more accurate path embeddings. In short, given the embeddings of each behavior path and its corresponding click behavior, PEM first activates the important behaviors in a behavior path and then optimizes the embedding of the behavior path so as to obtain a more accurate behavior path representation. 

Then, PMM uses the embeddings of the current behavior path and historical behavior paths as input to search the $k$ most similar historical behavior paths compared to the current behavior path and then activate the corresponding $k$ click behaviors with the candidate. 

In addition, PAM aims to learn more precise and informative behavior path embeddings by contrastive learning. Concretely, we mask each historical behavior path to obtain two augmented paths and feed them into the embedding layer and PEM to calculate their embeddings and then pull embeddings stemming from a same behavior path close by taking the InfoNCE loss \cite{oord2018InfoNCE} as the contrastive loss. 

DBPMaN uses the negative log-likelihood function as the main loss, which is widely used in most CTR models. 
Finally, DBPMaN is trained by combining the main loss with the contrastive loss as the optimization objective.

Due to the limited space, we only describe the PEM and PMM in detail, omitting the other parts of the model.

\subsection{Path Enhancing Module (PEM)}  
For a behavior path $[b_{m(i)-l},..., b_{m(i)-2}, b_{m(i)-1}]$ and the following click behavior $b_i^c$, the embedding layer will generate their embeddings, denoted by $\mathbf{e}_{m(i)-l}$,$...$, $\mathbf{e}_{m(i)-2}$, $\mathbf{e}_{m(i)-1}$, and $\mathbf{e}_i^c$. Further, a sequence of embeddings $[\mathbf{e}_{m(i)-l},...,\mathbf{e}_{m(i)-2}, \mathbf{e}_{m(i)-1}]$ is denoted by $\mathbf{s}_i$. 

We firstly apply a local activation unit on the user behavior path, which performs a weighted concat pooling to adaptively calculate the embedding of the behavior path, as shown in Eq. \ref{din_behavior_path}.

\vspace{-5pt}
\begin{equation}
\begin{aligned}
\mathbf{e}_{m(i)-j}^{te} &= a\left(\mathbf{e}_{m(i)-j}, \mathbf{e}_{i}^c\right)\cdot \mathbf{e}_{m(i)-j},\quad 1\le j\le l\\
\mathbf{s}_i^{te} &= [\mathbf{e}_{m(i)-l}^{te},...,\mathbf{e}_{m(i)-2}^{te},\mathbf{e}_{m(i)-1}^{te}]\\
&\mathbf{p}_i^{te} = \operatorname{concat}(\mathbf{s}_i^{te})\\
\end{aligned}
\label{din_behavior_path}
\end{equation}
\noindent where $a(\cdot)$ is an MLP whose output is used as the first-level activation score.

Then, we feed $\mathbf{p}_i^{te}$ into another MLP and learn the second-level activation score of each behavior in the behavior path by a softmax activation function, as shown in Eq. \ref{mask_score}.

\vspace{-5pt}
\begin{equation}
\mathbf{score}_i = \operatorname{softmax}(\operatorname{MLP}(\mathbf{p}_{i}^{te}))
\label{mask_score}
\end{equation}

The $\mathbf{score}_i$ is an $l$-dimensional vector, whose entries represent the second-level activation scores of behaviors in the behavior path. Then, only the top-k scores are chosen. According to the chosen scores, we multiply them with their corresponding behavior embeddings in $\mathbf{s}_i^{te}$. By concatenating the embeddings scaled by scores, we get the enhanced path embedding $\mathbf{p}_{i}^e$.

In this way, we can obtain the sequence of the enhanced path embeddings $\mathbf{P}_e = [\mathbf{p}_1^e,\mathbf{p}_2^e,...,\mathbf{p}_t^e]$.

\subsection{Path Matching Module (PMM)}  
For a user, there might exist a large number of behavior paths in the user behavior sequence. However, only a few of them are similar to the current behavior path, which can indicate the user's current interests. PMM is designed to search the first $k$ behavior paths most similar to the current path, and then obtain the corresponding $k$ click behaviors, which are believed to make considerable contributions to the user's current interests.

Specifically, given the sequence of the enhanced historical path embeddings $\mathbf{P}_e = [\mathbf{p}_1^e,\mathbf{p}_2^e,...,\mathbf{p}_t^e]$ and the enhanced embedding of the current behavior path $\mathbf{p}_{cur}^e$, we feed each $\mathbf{p}_i^e \in \mathbf{P}_e$ and $\mathbf{p}_{cur}^e$ into a scoring gate and obtain a similarity score $g^p_i$, which reflects the importance of the corresponding historical behavior path. The calculation of the scoring gate is shown in Eq. \ref{att_score}.

\vspace{-5pt}
\begin{equation}
g_i^p = \operatorname{MLP}(\operatorname{concat}(\mathbf{p}_{cur}^e,\mathbf{p}_i^e,\mathbf{p}_{cur}^e\otimes \mathbf{p}_i^e))
\label{att_score}
\end{equation}

\noindent where $\otimes$ denotes the hadamard product and $\operatorname{MLP}(\cdot)$ is implemented as a feed forward neural network. 

Thus, we can obtain a list of similarity scores $g^p = [g_1^p, g_2^p,...,g_t^p]$. We sort all scores and choose the top-k scores. With top-k scores, we can get the corresponding historical paths and the click behaviors corresponding to the historical paths. For the chosen click behaviors, a sequence of these click behavior embeddings is denoted by $\mathbf{s}_c = [\mathbf{e}_{c1}, \mathbf{e}_{c2},...,\mathbf{e}_{ck}]$. In addition, we multiply each chosen path embedding by the corresponding score and obtain the adjusted embeddings of chosen paths, as shown in Eq. \ref{filter}.

\vspace{-5pt}
\begin{equation}
\mathbf{E}^p = \operatorname{concat}(Filter(g^p, [g^p_i\cdot \mathbf{p}^e_i,1\le i\le t],k))
\label{filter}
\end{equation}

\noindent where the function $Filter(score, embedding,k)$ sorts historical behavior paths by score and chooses top-k paths. 

All chosen click behaviors are supposed to make different contributions to the user's current interests. Thus, we use the same way as one in Eq. \ref{att_score} to calculate the scores of similarity between the candidate and the chosen click behaviors so as to adaptively calculate the representation vector of user interests by taking into consideration the relevance between them, as shown in Eq. \ref{att_score2_and_att_sum}. 

\vspace{-5pt}
\begin{equation}
\begin{aligned}
g_i^c = \operatorname{MLP}(\operatorname{concat}(\mathbf{e}^{ct},\mathbf{e}_{ci},\mathbf{e}^{ct}\otimes \mathbf{e}_{ci}))\\
\mathbf{E}^c = \operatorname{concat}([g_i^c\cdot \mathbf{e}_{ci}, 1\le i\le k])
\label{att_score2_and_att_sum}
\end{aligned}
\end{equation}

At last, $\mathbf{P}_e$, $\mathbf{E}^p$, $\mathbf{E}^c$, $\mathbf{e}^u$ and $\mathbf{e}^{ct}$ are concatenated and then fed into an MLP layer which outputs the predicted CTR.




\section{Experimental Evaluation}
\subsection{Experimental Setup}
\textbf{Datasets.} We adopt the following two datasets for experiments.

\begin{itemize}[leftmargin=*]
\item[$\bullet$] Taobao: a public dataset \cite{zhu2018taobao_dataset}, containing 10-day interactions.
We preprocess the data in the same way as what CAN do in \cite{bian2022can}.  
\item[$\bullet$] Meituan: an industrial dataset collected by the Meituan Takeout App, which contains 14-day interactions of 100 million users.
\end{itemize}

Table\ref{tab:dataset} lists the statistics of processed datasets.

\noindent \textbf{Competitors.} We choose the following CTR models which focus on feature interaction modeling as comparison models. 

\begin{itemize}[leftmargin=*]
\item[$\bullet$] DeepFM \cite{guo2017deepfm}. It combines the factorization machines and deep learning for low-order and high-order feature interactions.
\item[$\bullet$] xDeepFM \cite{lian2018xdeepfm}. It generates feature interactions using the proposed Compressed Interaction Network (CIN) and further combines a CIN and a basic DNN into one unified model.
\item[$\bullet$] DIN \cite{zhou2018din}. It designs a local activation unit to learn the representation of user interests from historical behaviors w.r.t. a candidate.
\item[$\bullet$] DIEN \cite{zhou2019dien}. It designs an interest extractor layer and an interest evolving layer to capture interests from behavior sequences.
\item[$\bullet$] ONN \cite{yang2020onn}. It learns different representations for different operations.
\item[$\bullet$] CAN \cite{bian2022can}. It disentangles the representation learning and feature interaction modeling via the co-action unit. 
\end{itemize}

\noindent \textbf{Metrics.} We use AUC and RelaImpr \cite{yan2014RelaImpr} as the metrics in offline experiments, CTR and CPM (Cost-Per-Mille) as metrics in online experiments.

\noindent \textbf{Implementation Details.} We implement DBPMaN \footnote{The code is available at https://github.com/Ethan-Yys/DBPMaN.} by Tensorflow. For all models, we use Adam as the optimizer with a learning rate of 0.001. The model parameters are initialized with a Gaussian distribution (with a mean of 0 and a standard deviation of 0.01). 
The item embedding dimension is set to 18. 

\subsection{Performance Comparison}
We conduct comparative experiments, comparing our model with the above competitors. The performance results of different models on two datasets are shown in Table \ref{tab:performance results on offline evaluations}. 

From the results, we find the performance ranking of all the models over two datasets is the same and our DBPMaN surpasses all the competitors. We think that the attention mechanism on path-to-path activation in DBPMaN helps defeat the other models, including attention-based models (i.e., DIN, DIEN, and ONN) which employ point-to-point activation of a candidate and individual behavior in the historical behavior sequence. 

An interesting finding is that our DBPMaN model achieves more significant improvement over other models on the Meituan dataset rather than the Taobao dataset. This may stem from the nature of the food takeout scenario. The relatively few food stores surrounding the user, combined with characteristics of user interest in food, lead to repetitive interactions with the same food store in the history of user behaviors. This makes it easy to activate historical paths relevant to the current path.

\begin{table}[t]
\renewcommand{\arraystretch}{1.2}
  \setlength{\abovecaptionskip}{0cm} 
  \setlength{\belowcaptionskip}{-0.1cm}
  \small
  \caption{Statistics of datasets.}
  \label{tab:dataset}
  \begin{tabular}{lcccc}
    \toprule
    Datasets&\#Users&\#Items&\#Categories&\#Interactions\\
    \midrule
    Taobao & 987991 & 4161138 & 9437 & 100095182\\
    Meituan & 100000000 & 15755909 & 184 & 5648932411\\
  \bottomrule
\end{tabular}
\end{table}

\begin{table}
\renewcommand{\arraystretch}{1.2}
  \setlength{\abovecaptionskip}{0cm} 
  \setlength{\belowcaptionskip}{-0.1cm}
  \small
  \caption{Performance results on offline evaluations.}
  \label{tab:performance results on offline evaluations}
  \begin{tabular}{lcccc}
    \toprule 
    \multirow{2}{*}{Model} & \multicolumn{2}{c}{Taobao} & \multicolumn{2}{c}{Meituan} \\
    \cline{2-5}
    & AUC & RelaImpr & AUC & RelaImpr \\
    \midrule
    DeepFM & 0.8125& -13.19\% & 0.6673 &-1.55\%\\
    xDeepFM & 0.8366&-10.61\% & 0.6693 &-1.25\%\\
    ONN & 0.8689&-7.16\% & 0.6705 &-1.08\%\\
    DIN & 0.9308&-0.54\% & 0.6753 &-0.37\%\\
    DIEN & 0.9324&-0.37\% & 0.6761 &-0.25\%\\
    CAN & 0.9359&0.00\% & 0.6778 & 0.00\%\\
    \textbf{DBPMaN} & \textbf{0.9381} & \textbf{0.24\%} & 
    \textbf{0.6812} & \textbf{0.50\%}\\
  \bottomrule
\end{tabular}
\end{table}

\subsection{Ablation Study}
We conduct an ablation study on the Meituan dataset to evaluate the contributions of key modules of DBPMaN. We compare our model with three variants, i.e., DBPMaN w/o PEM, DBPMaN w/o PMM, and DBPMaN w/o PAM.
The results are shown in Table \ref{tab:ablation study on Meituan dataset}.

From Table \ref{tab:ablation study on Meituan dataset}, we can find that three variants suffer a decrease in all three metrics, compared to the original DBPMaN. The performance of DBPMaN w/o PMM declines the most,
which indicates that PMM plays a more important role than the other two modules. 
 
\begin{table}
\renewcommand{\arraystretch}{1.2}
  \setlength{\abovecaptionskip}{0cm} 
  \setlength{\belowcaptionskip}{-0.1cm}
  \small
  \caption{Ablation study on Meituan dataset.}
  \label{tab:ablation study on Meituan dataset}
  \begin{tabular}{lccc}
    \toprule
    Model&AUC&CTR&CPM\\
    \midrule
    DBPMaN & 0.6812 & - & - \\ 
    DBPMaN w/o PEM & 0.6789 & -0.6\% & -0.5\% \\
    DBPMaN w/o PMM & 0.6769 & -1.0\% & -1.1\% \\
    DBPMaN w/o PAM & 0.6801 & -0.2\% & -0.2\% \\
  \bottomrule
\end{tabular}
\end{table}

\subsection{Online A/B test}


The A/B test is conducted on the Meituan food delivery platform and lasts for 14 days from 2022-08-10 to 2022-08-23, where the baseline model is our last online CTR model which only uses the point-to-point activation method. The results are shown in Table \ref{tab:ab_test}, where $l$ denotes the length of the behavior path. Now DBPMaN ($l$=8) has been deployed online and serves the main traffic of users.




\begin{table}
\renewcommand{\arraystretch}{1.2}
  \setlength{\abovecaptionskip}{0cm} 
  \setlength{\belowcaptionskip}{-0.1cm}
  \small
  \caption{Online A/B test results.}
  \label{tab:ab_test}
  \begin{tabular}{lccc}
    \toprule
    Model&CTR&CPM&Inference Time(ms)\\
    \midrule
    Baseline & - & - & 48\\
    DBPMaN ($l$=8) & +1.6\% &+1.8\%&+0.8\\
    DBPMaN ($l$=16) & +1.3\% &+1.5\%&+2.1\\
  \bottomrule
\end{tabular}
\vspace{-10pt}
\end{table}

\section{Conclusion}
In this paper, we propose DBPMaN, which models user behavior paths into CTR prediction for the first time. Besides the excellent performance, DBPMaN shows the possibility of exploring user decision-making psychology by modeling behavior paths. 

\bibliographystyle{ACM-Reference-Format}
\bibliography{sample-base}

\end{document}